\documentclass{article}

\usepackage{graphicx}
\usepackage{subfigure}

\usepackage[utf8]{inputenc} 
\usepackage[T1]{fontenc}    

\usepackage{hyperref}       
\hypersetup{colorlinks, citecolor=[rgb]{0,0.08,0.65}, urlcolor=[rgb]{0,0.08,0.65}, linkcolor=[rgb]{0,0.08,0.65}, filecolor=[rgb]{0,0.08,0.65}}

\usepackage{url}            
\usepackage{booktabs}       
\usepackage{bm}
\usepackage{dsfont}
\usepackage{amsfonts}       
\usepackage{nicefrac}       
\usepackage{microtype}      
\usepackage{amsmath}
\usepackage{amsthm}
\usepackage{amssymb}
\usepackage{amsfonts}
\usepackage{algorithm,algorithmic}
\usepackage{caption}

\usepackage{hyperref}

\usepackage{amsmath,amsfonts,bm}

\def\eqref#1{equation~\ref{#1}}

\def\1{\bm{1}}

\def\vmu{{\bm{\mu}}}

\def\vphi{{\bm{\phi}}}
\def\vsigma{{\bm{\sigma}}}

\def\vh{{\bm{h}}}

\def\vx{{\bm{x}}}
\def\vy{{\bm{y}}}

\def\mW{{\bm{W}}}
\def\mX{{\bm{X}}}

\def\mLambda{{\bm{\Lambda}}}
\def\mXi{{\bm{\Xi}}}
\def\mSigma{{\bm{\Sigma}}}

\DeclareMathAlphabet{\mathsfit}{\encodingdefault}{\sfdefault}{m}{sl}
\SetMathAlphabet{\mathsfit}{bold}{\encodingdefault}{\sfdefault}{bx}{n}

\usepackage[accepted]{icml2019}

\icmltitlerunning{Dropout as a Structured Shrinkage Prior}

\begin{document}

\twocolumn[
\icmltitle{Dropout as a Structured Shrinkage Prior}



\icmlsetsymbol{equal}{*}

\begin{icmlauthorlist}
\icmlauthor{Eric Nalisnick}{cam}
\icmlauthor{Jos\'e Miguel Hern\'andez-Lobato}{cam,micro,turing}
\icmlauthor{Padhraic Smyth}{uci}
\end{icmlauthorlist}

\icmlaffiliation{cam}{Department of Engineering, University of Cambridge, Cambridge, United Kingdom}
\icmlaffiliation{micro}{Microsoft Research, Cambridge, United Kingdom}
\icmlaffiliation{turing}{Alan Turing Institute}
\icmlaffiliation{uci}{Department of Computer Science, University of California, Irvine, United States of America}

\icmlcorrespondingauthor{Eric Nalisnick}{e.nalisnick@eng.cam.ac.uk}

\icmlkeywords{Bayesian, Dropout, Shrinkage}

\vskip 0.3in
]



\printAffiliationsAndNotice{}  

\begin{abstract}
Dropout regularization of deep neural networks has been a mysterious yet effective tool to prevent overfitting.  Explanations for its success range from the prevention of "co-adapted" weights to it being a form of cheap Bayesian inference.  We propose a novel framework for understanding multiplicative noise in neural networks, considering continuous distributions as well as Bernoulli noise (i.e.\ dropout).  We show that multiplicative noise induces structured shrinkage priors on a network's weights.  We derive the equivalence through reparametrization properties of scale mixtures and without invoking any approximations.  Given the equivalence, we then show that dropout's Monte Carlo training objective approximates marginal MAP estimation.  We leverage these insights to propose a novel shrinkage framework for resnets, terming the prior \textit{automatic depth determination} as it is the natural analog of \textit{automatic relevance determination} for network depth.  Lastly, we investigate two inference strategies that improve upon the aforementioned MAP approximation in regression benchmarks.
\end{abstract}

\section{Introduction}
\textit{Dropout} regularization \citep{hinton2012improving, srivastava2014dropout} has become an essential tool for fitting large neural networks.  Due to its success, a number of variants have been proposed \citep{wan2013regularization, wang2013fast, huang2016deep, singh2016swapout, achille2018information}, including versions for recurrent \citep{ji2015blackout, krueger2016zoneout, gal2016theoretically, zolna2018fraternal} and convolutional \citep{tompson2015efficient, gal2016conv} architectures.  The narratives attempting to explain dropout's inner-workings and success are also plentiful.  To give a few examples, \citet{srivastava2014dropout} argue it prevents "conspiracies" between hidden units, \citet{hinton2012improving} claim it serves a role similar to sex in evolution, \citet{baldi2013understanding} show it ensembles by taking the geometric mean of sub-models, \citet{wager2013dropout} explain it as an adaptive ridge penalty, and \citet{gal2016dropout} suggest dropout performs quasi-Bayesian uncertainty estimation.  While some prior work has shown strict equivalences for simple models such as linear regression \citep{baldi2013understanding, wager2013dropout, wager2014altitude, helmbold2015inductive}, the general case of dropout in deep neural networks is analytically intractable, which is likely why no one narrative has come to prominence.

In this paper we propose a novel Bayesian interpretation of regularization via multiplicative noise---with dropout being the special case of Bernoulli noise.  Unlike previous frameworks, our method of analysis works through reparametrizations that are agnostic to network architecture.  By assuming nothing more than a Gaussian prior (which could be diffuse) on the weights, we show that multiplicative noise induces (marginally) a Gaussian scale mixture.  This result is exact and has been exploited previously in Bayesian modeling \citep{kuo1998variable}.  Not only do we lay bare multiplicative noise's distributional assumptions, but we also reveal the structure it induces on the network's weights.  We find that noise applied to hidden units ties the scale parameters in the same way as \textit{automatic relevance determination} \citep{neal1995bayesian, mackay1996bayesian, tipping2001sparse}, a well-studied shrinkage prior.  We propose an extension of this prior for residual networks \citep{he2016Deep}, allowing Bayesian inference to select the number of layers.

We also address our framework's implications for posterior inference.  We show that dropout's Monte Carlo objective is a lower bound on the scale mixture model's marginal MAP objective.  Decoupling dropout's model from inference is a novel and useful contribution as previous Bayesian interpretations have been grounded in variational inference \citep{gal2016dropout, kingma2015variational} and hence gave no guidance on how dropout could be used in conjunction with Markov chain Monte Carlo (MCMC).  We then make algorithmic contributions of our own, describing a computationally efficient importance weighted objective and EM algorithm.  We test these algorithms on benchmark regression tasks from the UCI repository \citep{Dua:2017}, showing our proposals for light-weight inference improve upon traditional Monte Carlo dropout and are competitive to other high-capacity Bayesian models such as deep Gaussian processes trained with expectation propagation.  Lastly, we leave the reader with some directions for future work.    

\section{Background}
We use the following notation throughout the paper.  Matrices are denoted with upper-case and bold letters (e.g.\ $\mX$), vectors with lower-case and bold (e.g.\ $\vx$), and scalars with no bolding (e.g.\ $x$ or $X$).  Data are assumed to be row vectors $\vx \in \mathbb{R}^{D}$, and $N$ independently and identically distributed observations constitute the empirical data set $\mX = \{\vx_{1}, \ldots,  \vx_{N}\}$.  We focus on supervised learning tasks in which $\mX$ are covariates (features) that are predictive of another variable $\vy = \{ y_{1}, \ldots, y_{N} \}$, which we assume is a one-dimensional regression response or index denoting a class label.  Throughout we use $r$ to index the rows of a matrix and $j$ to index its columns.  We define an $L$-layer \textit{neural network} (NN) \citep{goodfellow2016Deep} recursively as \begin{equation}\label{layer_compute}\begin{split}
      &\mathbb{E}[y_{n} | \vx_{n}] = g^{-1}( \vh_{n,L} \mW_{L+1}), \\ &\vh_{n,l} = f_{l}( \vh_{n, l-1}  \mW_{l}),  \ \ \ \ \vh_{n,0} = \vx_{n}
\end{split}\end{equation} where $g(\cdot)$ is a link function following the GLM framework \citep{nelder2004generalized}. $\{\mW_{1},\ldots,\mW_{l},\ldots,\mW_{L+1}\}$ are the parameters, a set of $D_{l-1} \times D_{l}$-dimensional matrices.  We omit the bias terms to reduce notational clutter as they can be subsumed into the weight matrices.  The function $f(\cdot)$ acts element-wise and is known as the \textit{activation} function.  

\paragraph{Multiplicative Noise Regularization (Dropout)}  Dropout training \citep{hinton2012improving, srivastava2014dropout} introduces \textit{multiplicative noise} (MN) into the hidden layer computation defined in Equation \ref{layer_compute}: \begin{equation}\label{dropout_compute}
\vh_{n,l} = f_{l}(\vh_{n,l-1}\mLambda_{l}\mW_{l})
\end{equation} where $\mLambda_{l}$ is a diagonal $D_{l-1} \times D_{l-1}$-dimensional matrix of random variables $\lambda_{j,j}$ drawn independently from a noise distribution $p(\lambda)$.  Dropout corresponds to $p(\lambda)$ being Bernoulli.  However, other noise distributions such as Gaussian \citep{srivastava2014dropout, contdropout}, Beta \citep{tomczak2013prediction, liu2019beta}, and uniform \citep{contdropout} have been shown to be equally effective.  Training under MN is done by sampling a new $\mLambda_{l}$ matrix for every forward propagation.  This sampling can be viewed as Monte Carlo (MC) integration over the noise distribution, and therefore, the MN optimization objective is to maximize w.r.t $\{ \mW_{l} \}_{l=1}^{L+1}$ the function \begin{equation}\label{dropout_loss}\begin{split}
    \mathcal{L}_{\text{MN}}&(\{ \mW_{l} \}_{l=1}^{L+1}) \\ &= \mathbb{E}_{p(\lambda)}[\log p(\vy | \mX, \{ \mW_{l} \}_{l=1}^{L+1}, \{\mLambda_{l}\}_{l=1}^{L})] \\ &\approx \frac{1}{S} \sum_{s=1}^{S} \log p(\vy | \mX, \{ \mW_{l} \}_{l=1}^{L+1},\{\hat{\mLambda}_{l,s}\}_{l=1}^{L})
\end{split}
\end{equation} where the expectation is taken with respect to $p(\lambda)$ and $\hat{\mLambda}_{l, s}$ denotes the $s$th set of samples for the $l$th layer.  

\paragraph{Automatic Relevance Determination}  \textit{Automatic relevance determination} (ARD) \citep{mackay1996bayesian, neal1995bayesian, tipping2001sparse} is a Bayesian regularization framework that consists of placing (usually) Gaussian priors on the NN's weights and then structured hyper-priors on the Gaussian scales.  The scales of the weights in the same row (assuming the matrix orientation in Equation \ref{layer_compute}) are tied so they grow or shrink together in a form of group regularization.  The end result is feature / hidden unit selection since if all of a unit's outgoing weights are near zero, then the unit is inconsequential to the model output.  We can write the ARD prior as \begin{equation}\label{ardEq}
     w_{l,r, j} \sim \text{N}(0, \sigma_{l,r}^{2}), \ \ \ \sigma_{l,r} \sim p(\sigma) \end{equation} 
where $l$ is the index on layers, $r$ is the index on rows in the weight matrix, and $j$ is the index on its columns.  Writing $\sigma_{l,r}$ without a column index signifies that all of the weights in the $r$th row share the same scale.  Although we have defined ARD using a first-level Gaussian prior, other distributions could be used as long as they can be given the same scale structure.

\section{Multiplicative Noise as Automatic Relevance Determination}\label{sec:MNasARD}
We now discuss our first contribution: showing that regularization via MN induces, under mild assumptions, an ARD prior.  The key observation is that if we assume the weights to be Gaussian random variables, the product $\mLambda_{l} \mW_{l}$ defines a Gaussian scale mixture (GSM) with ARD structure (\textit{ARD-GSM prior} for short).  We then show how the MC training objective in Equation \ref{dropout_loss} can be derived from this framework.  

\subsection{Gaussian Scale Mixtures} A random variable $\theta$ is a \textit{Gaussian scale mixture} (GSM) if (and only if) it can be expressed as the product of a Gaussian random variable--call it $z$--with zero mean and some variance $\sigma_{0}^{2}$ and an independent scalar random variable $\alpha$ \citep{beale1959scale, andrews1974scale}:
\begin{equation}\label{GSM_def}
\theta \,{\buildrel d \over =}\, \alpha z, \ \ \ z \sim \text{N}(0, \sigma_{0}^{2}), \ \ \alpha \sim p(\alpha) 
\end{equation} where $\,{\buildrel d \over =}\,$ denotes equality in distribution.  The RHS is known as the GSM's \textit{expanded parametrization} \citep{kuo1998variable}.  While it may not be obvious from Equation \ref{GSM_def} that $\theta$ is a \textit{scale} mixture, the result follows from the Gaussian's closure under linear transformations: $\alpha z \sim \text{N}(\alpha \cdot 0,  \alpha^{2} \cdot \sigma^{2}_{0})$.  Integrating out the scale gives the marginal density of $\theta$: 
$p(\theta) = \int \text{N}(0,\sigma_{0}^{2} \alpha^{2}) p(\alpha) d\alpha$ where $p(\alpha)$ is now clearly the mixing distribution.  We call the form $\text{N}(w; 0,\sigma_{0}^{2} \alpha^{2}) p(\alpha)$ the \textit{hierarchical parametrization}.  Super-Gaussian distributions, such as the student-t, Laplace, and horseshoe \citep{carvalho2009handling}, can be represented as GSMs, and the hierarchical parametrization is often used for its convenience when employing them as robust priors \citep{steel2000bayesian}.

\subsection{Equivalence Between MN and ARD-GSM Priors} Now that we have defined GSMs, we demonstrate their relationship to MN.  Assume we have an $L$-layer Bayesian NN with an ARD-GSM prior: \begin{equation}\begin{split}\label{gsm1}
    &y_{n} \sim p(y_{n} | \vx_{n}, \{ \mW_{l} \}_{l=1}^{L+1}) \\ &w_{l,r,j} \sim \text{N}(0, \sigma^{2}_{0} \xi^{2}_{l,r}), \ \ \xi_{l,r} \sim p(\xi)  
\end{split}\end{equation} where $w_{l,r,j}$ denotes the NN weights, $\sigma^{2}_{0}$ a constant shared across all weights, and $\xi_{l, r}$ a local (row-wise) random scale.  Accordingly we have given $\xi_{l, r}$ a layer index ($l$) and row index ($r$) but not a column index ($j$), following Equation \ref{ardEq}.  

As the prior on the weights is a GSM, we can reparametrize the model into the GSM's equivalent expanded form given in Equation \ref{GSM_def}: \begin{equation}\begin{split}
    &y_{n} \sim p(y_{n} | \vx_{n}, \{ \mW_{l} \}_{l=1}^{L+1}, \{\mXi_{l}\}_{l=1}^{L}) \\ &w_{l,r,j} \sim \text{N}(0, \sigma_{0}^{2}), \ \ \xi_{l, r} \sim p(\xi)  
\end{split}\end{equation} where the weights are still denoted $w_{l,r,j}$ and drawn from a Gaussian with a fixed variance.  The reparametrization changes the NN's hidden layer computation to: \begin{equation}\begin{split}\label{reparam1}
\vh_{n,l}  = f_{l}(\vh_{n,l-1}\mW_{l})  \,{\buildrel \text{reparametrization} \over \longrightarrow}\, \  f_{l}(\vh_{n,l-1}\mXi_{l}\mW_{l}) 
\end{split}\end{equation} where $\mXi_{l}$ is a diagonal $D_{l-1}\times D_{l-1}$-dimensional matrix of scale values $\xi_{l,r}$.  Notice that this expression (Equation \ref{reparam1}) is Equation \ref{dropout_compute} exactly---with $\xi$ in the place of $\lambda$---and thus we have shown the equivalence to MN.  We have used no approximations.  Moreover, the Gaussian assumption is not a strong one, and it can be relaxed by making $\sigma_{0}^{2}$ sufficiently large so that the prior is diffuse and therefore negligible.  Previous attempts at analyzing MN have been frustrated by the activation function, which introduces a composition of non-linear functions that makes expectations analytically intractable.  Because the expanded-vs-hierarchical reparametrization acts through the inner products, the activation function is bypassed.

\subsection{Monte Carlo Training as Marginal MAP Inference}\label{margMAPsec}
Next we derive the dropout / MN optimization objective given in Equation \ref{dropout_loss} from the ARD-GSM perspective.  Specifically, $\mathcal{L}_{\text{MN}}(\{ \mW_{l} \}_{l=1}^{L+1})$ is equivalent to a lower bound on the marginal MAP objective---i.e.\ the objective that would be optimized to find the weights' MAP estimate (assuming the optimum could be found).  We begin by writing the unnormalized, marginal log-posterior over the weights as 
\begin{equation}\begin{split} &\log p( \{ \mW_{l} \}_{l=1}^{L+1} | \vy, \mX) \\  
& \propto \log \mathbb{E}_{p(\xi)}\left[ p(\vy | \mX, \{ \mW_{l} \}_{l=1}^{L+1}, \{\mXi_{l}\}_{l=1}^{L}) \right]  \\ & \ \ \ \ \ \ \ \ \ \ \ \ \ \ \ \ \ \ \ \ \ \ \ \ \ \ \ \ \ \ \ \ \ \ \ \ \ \ \ \ + \frac{-1}{2\sigma^{2}_{0}} \sum_{l=1}^{L+1} \sum_{r=1}^{D_{l-1}} \sum_{j=1}^{D_{l}} w_{l,r,j}^{2} \\ &= \mathcal{L}_{\text{MAP}}(\{ \mW_{l} \}_{l=1}^{L+1}). 
\end{split}\end{equation}  Next we use Jensen's inequality to lower-bound the first term (i.e. $\log \mathbb{E}_{p(\xi)}[ \cdot ] \ge  \mathbb{E}_{p(\xi)}[ \log \cdot ]$): \begin{equation}\label{gsmFinal}\begin{split} & \mathcal{L}_{\text{MAP}} \ge \mathbb{E}_{p(\xi)} \left[ \log  p(\vy | \mX, \{ \mW_{l} \}_{l=1}^{L+1}, \{\mXi_{l}\}_{l=1}^{L}) \right] \\ & \ \ \ \ \ \ \ \ \ \ \ \ \ \ \ \ \ \ \ \ \ \ \ \ \ \ \ \ \ \ \ \ \ \ \ \ \ \ \ \    + \frac{-1}{2\sigma^{2}_{0}} \sum_{l=1}^{L+1} \sum_{r=1}^{D_{l-1}} \sum_{j=1}^{D_{l}} w_{l,r,j}^{2} 
\\ &= \mathcal{L}_{\text{MN}}(\{ \mW_{l} \}_{l=1}^{L+1}) +  \frac{-1}{2\sigma^{2}_{0}} \sum_{l=1}^{L+1} || \mW_{l} ||_{F}^{2}
\end{split}\end{equation} where $\mathcal{L}_{\text{MN}}$ is the MN objective defined in Equation \ref{dropout_loss} and $|| \cdot ||_{F}$ is the Frobenius norm.  Thus, the lower bound is equivalent to the MN objective with an additional L2 penalty (a.k.a.\ weight decay).  Using weight decay and MN regularization together is not uncommon; \citet{srivastava2014dropout} (see their Table 9) and \citet{gal2016dropout} both do so.  The L2 term can be removed by assuming that the Gaussian prior is sufficiently diffuse: $$\frac{-1}{2\sigma^{2}_{0}} \sum_{l=1}^{L+1} || \mW_{l} ||_{F}^{2} \rightarrow 0 \ \ \ \ \text{ as } \ \ \ \  \sigma^{2}_{0} \rightarrow \infty.$$  Increasing the Gaussian's variance requires adapting $p(\xi)$ to ensure the same shrinkage level but presents no technical difficulty otherwise.  From here forward we use $\xi$ to denote both MN and random scales and use $\lambda$ on its own to denote MN schemes proposed in prior work.


\subsection{Corresponding Priors} Having shown the equivalence between ARD-GSM priors and MN, we now discuss some specific noise distributions and their corresponding priors.  Starting with dropout, the noise distribution is $\xi \sim \text{Bernoulli}(\pi)$, and this implies the prior on the Gaussian's variance is also Bernoulli, i.e. $\xi^{2} \sim \text{Bernoulli}(\pi)$, since the square of a Bernoulli random variable is still a Bernoulli of the same distribution.  The marginal prior on the NN weights is then \begin{equation}\begin{split}
    p_{\text{\tiny{DROPOUT}}}(w) &= \sum_{\xi \in \{0, 1\}}  \ \text{N}(w; 0, \xi \sigma_{0}^{2}) \  p(\xi) \\ &= \pi \ \text{N}(w; 0, \sigma_{0}^{2}) + (1-\pi) \ \delta[w]
\end{split}\end{equation} where $\delta[ \cdot]$ denotes the delta function located at zero.  This is the \textit{spike-and-slab} prior commonly used for Bayesian variable selection \citep{mitchell1988bayesian,george1993variable, kuo1998variable}.  Interestingly, the expanded parametrization was used for linear regression by \citet{kuo1998variable}, and thus their work should be considered a precursor to dropout.  However, \citet{kuo1998variable} were interested in obtaining the feature inclusion probabilities $p(\xi=1 | \vy, \mX)$ rather than deriving a regularization mechanism to improve predictive performance.  When dropout is performed without weight decay, its prior becomes $p_{\text{\tiny{DROP}}}(w) \propto  \pi \ \mathds{1} + (1-\pi) \ \delta[w]$ where the improper uniform distribution $\mathds{1}$ is derived by taking $\sigma_{0} \rightarrow \infty$.  

In Table \ref{gsmCorr}, we list several additional noise models, their corresponding priors on the Gaussian variance, and their marginal distribution on the NN weights.  Gaussian MN corresponds to a $\chi^{2}$-distribution on the variance and a generalized hyperbolic \citep{barndorff1977exponentially} marginal distribution.  Other notable cases are Rayleigh noise, which corresponds to a Laplace marginal, inverse Nakagami noise \citep{nakagami1960m}, which corresponds to a student-t, and half-Cauchy noise, which corresponds to the horseshoe prior \citep{carvalho2009handling}.  

\subsection{Equivalence to DropConnect} If we assume all weights have independent scales, thus removing the ARD structure, the hidden layer computation in the expanded parametrization changes to: $\vh_{n,l} = f_{l}(\vh_{n,l-1}(\mXi_{l} \odot \mW_{l}))$ where $\odot$ denotes an element-wise product and $\mXi_{l}$ is now a dense matrix of scale variables.  Following the same derivation from this point reveals an equivalence to \textit{DropConnect} regularization \citep{wan2013regularization}, which applies MN to each weight instead of each hidden unit.  This absence of the regularizing ARD structure may explain why DropConnect has not been used as widely as dropout. 

\begin{table}
\begin{center}
\begin{tabular}{  c | c | c  } 
  Noise Model & Variance Prior  & Marginal Prior  \\ $p(\xi)$ &  $p(\xi^{2})$  &  $p(w)$\\
\hline 
Bernoulli  & Bernoulli & Spike-and-Slab  \\
Gaussian & $\chi^{2}$ & Gen.\ Hyperbolic \\ 
Rayleigh & Exponential & Laplace \\ 
Inverse Nakagami & $\Gamma^{-1}$ & Student-t \\ 
Half-Cauchy & Unnamed & Horseshoe 
\end{tabular}
\caption{\textit{Noise Models and their Corresponding GSM Prior.}}
\label{gsmCorr}
\end{center}
\end{table}

\section{Extension to Resnets}\label{sec:resnets}

With the previous section's insights in mind, we turn our attention to other NN architectures.  \textit{Resnets} \citep{he2016Deep} are NNs with \textit{residual connections} (a.k.a.\ skip connections) \citep{lang1988resnet, he2016Deep, srivastava2015training} between their hidden layers.  Residual connections simply add the previous hidden state to the usual non-linear transformation: 
    $\vh_{l} = f_{l}( \vh_{l-1} \mW_{l}) + \vh_{l-1}$.
Since a residual connection allows information to bypass the non-linear transformation, entire weight matrices can be shrunk to zero without obstructing the NN's forward propagation.  Thus we can create a prior that selects for \emph{layers} by tying the variance of all weights in the same matrix.  By collectively shrinking all the weights in coordination, we can reduce the layer's influence, effectively pruning it in the case of absolute shrinkage to zero.  We term this prior \textit{automatic depth determination} (ADD) as it is the natural analog of ARD for network depth.  ADD is specified as \begin{equation}\label{ADD}
     w_{l,r,j} \sim \text{N}(0, \sigma_{0}^{2} \tau_{l}^{2}), \ \ \  \tau_{l} \sim p(\tau) \end{equation} where we have introduced the variable $\tau_{l}$ that acts as a \emph{per-layer} group variance.   We denote this structure by giving $\tau$ a layer index $l$ but not a row or column index.  As $p(\tau_{l})$ places more density near zero, Bayesian inference will increasingly prefer to prune whole weight matrices, i.e. $\mathbf{h}_{l} \approx  f_{l}( \vh_{l-1} \mathbf{0} ) + \vh_{l-1} = \vh_{l-1}$ (assuming $f_{l}$ is a ReLU), making the network effectively more shallow.  In the supplementary materials, we show how ADD can be formulated as MN, which reveals equivalences to \textit{stochastic depth} regularization \citep{huang2016deep}. 
     
\begin{figure*}
\subfigure[Distribution of Importance Weights]{\includegraphics[width=.47 \linewidth]{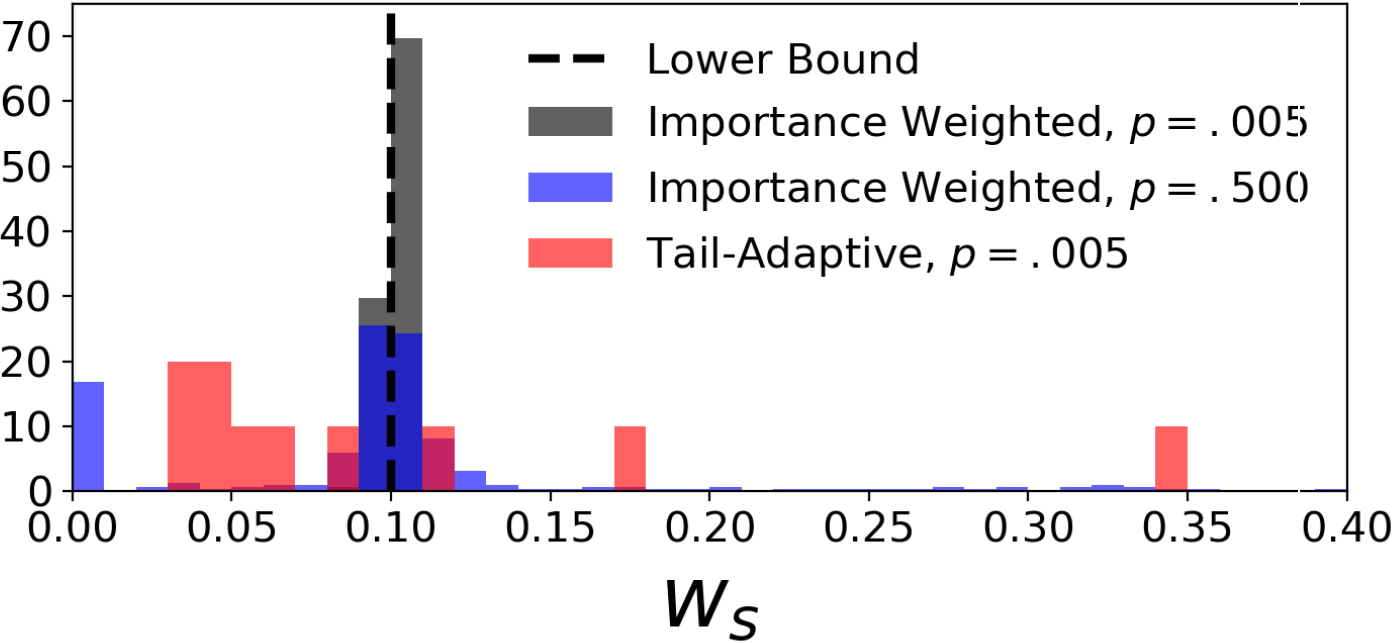} \label{fig:is_dist}}
\hfill
\subfigure[EM Updates for Variance]{\includegraphics[width=.49 \linewidth]{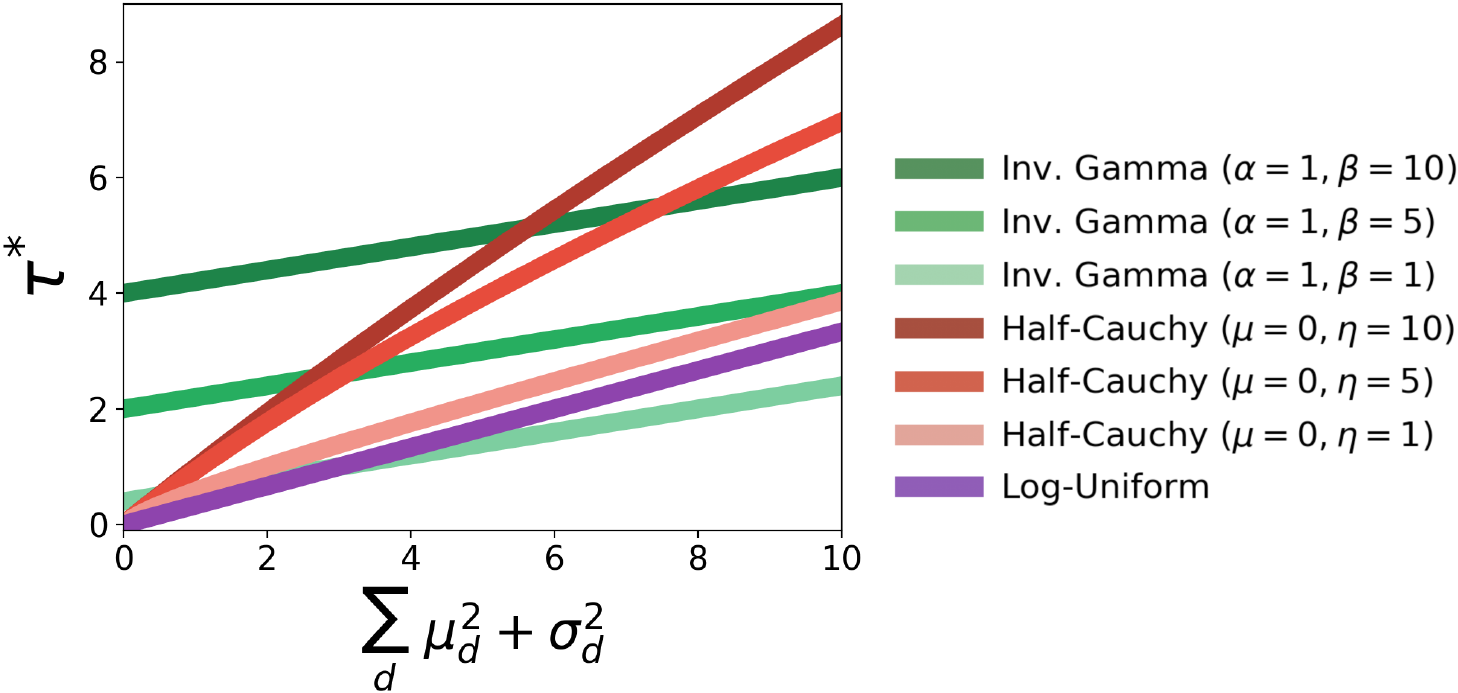}}
\caption{Subfigure (a) shows the empirical distribution of importance weights observed when training on \texttt{Energy}.  Subfigure (b) shows the EM updates for the posterior variance as a function of $q(\mW)$'s parameters. 
}
\label{fig:compare_w_vs}
\end{figure*}
 If Bayesian inference does not decide to prune an entire layer under the ADD prior, regularization still may be necessary.  The natural progression is to then select the layer's number of hidden units---as ARD does.  Therefore it makes sense to combine ARD and ADD so that the former takes effect when the latter imposes little to no regularization.  The joint ARD-ADD prior is specified as \begin{equation}\label{ardadd} w_{l,i,j} \sim \text{N}(w_{l,i,j}; 0, \sigma_{\small{0}}^{2} \xi_{l, r} \tau_{l}), \  \xi_{l, r} \sim p(\xi), \  \tau_{l} \sim p(\tau). \end{equation}  The priors remain essentially unchanged from their original definitions in Equations \ref{ardEq} and \ref{ADD}.  The two multiplicatively interact in the first-level prior's variance, and therefore when $\tau_{l} \rightarrow 0$, the product $\lambda_{l, i}\tau_{l} \approx 0$, effectively turning off the influence of ARD.  Conversely, when $\tau_{l} > 0$, then ARD will act as usual but have its effect modified by a factor of $\tau_{l}$.  Switching to the expanded parametrization, the hidden layer computation for ARD-ADD is  $ \vh_{n,l} = f_{l}(\tau_{l} \vh_{n,l-1} \mXi_{l} \mW_{l}) +  \vh_{n, l-1}.$  Implementing the prior as MN would involve sampling $\tau_{l}$ and $\mXi_{l}$ for each forward pass.

\section{Rethinking Dropout Inference} \label{sec:inf}
We next turn towards training and inference, speculating: are there ways to make dropout `more Bayesian'?  Can we optimize a bound that is closer to the true marginal map objective?  Or better yet, can we improve the posterior approximation while not incurring much additional cost in memory or computation?  In the next two subsections, we address these questions and propose two solutions.  Firstly, for traditional MC dropout we propose using a rank-based scheme for setting the importance weights, resulting in an objective that better covers posterior mass.  Secondly, for applications in which there is room to perform a little more computation, we detail a light-weight variational EM algorithm for ARD-type priors. 

\subsection{Expanded vs Hierarchical Parametrization}
Returning to Equation \ref{gsmFinal}, recall that the MC dropout objective is a lower bound on the true marginal MAP objective.  We are concerned with the gap between the two---
$\mathcal{J}_{\text{GAP}} = \mathcal{L}_{\text{MAP}} - \mathcal{L}_{\text{MN}}$---and wonder if it can be improved by changing parametrizations.  MN is always applied in the expanded parametrization, but we could alternatively apply noise in the hierarchical parametrization: \begin{equation}\label{hp_obj}\begin{split}
    \mathcal{L}_{\text{MN-HP}} = & \log p(\vy | \mX, \{ \mW_{l} \}_{l=1}^{L+1}) \\ & + \frac{-1}{2\sigma^{2}_{0}} \sum_{l=1}^{L+1} \sum_{r=1}^{D_{l-1}} \mathbb{E}_{p(\xi)} \left[ \frac{1}{\xi^{2}_{l,r}} \right] \sum_{j=1}^{D_{l}} w_{l,r,j}^{2}
\end{split}\end{equation} where the expectation would be computed with an MC approximation or solved for in closed-form.  At first glance the above objective would seem to be better behaved as the NN is no longer being perturbed.  However, in the supplementary materials we show that the expanded parametrization (i.e. likelihood noise) relaxes the MAP estimate as a function of $Var[\xi]$ whereas the alternative above has a $Var[\xi^{-2}]$-order gap.  When $\mathbb{E}[\xi^{2}]$ is near zero, $Var[\xi^{-2}]$ explodes for all noise distributions considered in Table \ref{gsmCorr}, resulting in an impractical objective.  The intuition behind these results is easy to see in the case of Bernoulli noise: multiplying the NN's weights by $\xi \in \{0, 1 \}$ is not a problem but plugging $\xi=0$ into Equation \ref{hp_obj} results in the second term becoming undefined.  The implication is that, of the two parametrizations considered, the expanded form provides the tightest bound to the true MAP objective under moderate to strong shrinkage. 

\subsection{Importance Weighted Objectives}\label{subsec:is} The following MC importance weighted objective is attractive as it is tighter than the $\mathcal{L}_{\text{MN}}$ lower bound \citep{burda2015importance, noh2017regularizing}: \begin{equation}\label{eq:is_obj}\begin{split}
\log & \ \mathbb{E}_{p(\xi)}[p(\vy | \mX, \{ \mW_{l} \}_{l=1}^{L+1}, \{\mXi_{l}\}_{l=1}^{L})] \\ &\ge  \log \frac{1}{S} \sum_{s=1}^{S} p(\vy | \mX, \{ \mW_{l} \}_{l=1}^{L+1}, \{\hat{\mXi}_{l,s}\}_{l=1}^{L}) \\ &= \mathcal{L}_{\text{IW-MN}}(\{ \mW_{l} \}_{l=1}^{L+1})
\end{split}\end{equation} where $\{\hat{\mXi}_{l,s}\}_{l=1}^{L}$ is the $s$th sample from $p(\xi)$.  While this objective is also a lower bound, it is guaranteed to become tighter with every additional sample, eventually converging to $\mathcal{L}_{\text{MAP}}$ as $S \rightarrow \infty$ \cite{burda2015importance, noh2017regularizing}.  The benefits of the IW objective can be seen in its gradient updates:      
\begin{equation}\label{iw_grad}\begin{split}
  & \nabla_{\mW} \mathcal{L}_{\text{IW-MN}}(\{ \mW_{l} \}_{l=1}^{L+1}) = \\ & \ \  \sum_{s=1}^{S} w_{s} \nabla_{\mW} \log p(\vy | \mX, \{ \mW_{l} \}_{l=1}^{L+1}, \{\hat{\mXi}_{l,s}\}_{l=1}^{L}) \end{split}\end{equation} where $\tilde{w}_{s} = p(\vy | \mX, \{ \mW_{l,s} \}_{l=1}^{L+1}, \{\hat{\mXi}_{l,s}\}_{l=1}^{L})$ and $w_{s} = \tilde{w}_{s} / \sum_{k} \tilde{w}_{k}$.  Samples that result in a higher likelihood exhibit more influence on the aggregated gradient.  \citet{noh2017regularizing} showed that this IW objective improves dropout's performance in several vision tasks.

\paragraph{Tail-Adaptive Weights} Yet, if the motivation for using dropout is to perform cheap uncertainty quantification \citep{gal2016dropout}, we should be optimizing an objective that covers as much posterior mass as possible.  This is even more desirable when using parameter point estimates as hopefully better posterior exploration will mitigate the effect of such a poor posterior approximation.  We propose using \citet{wang2018tailadaptive}'s \textit{tail-adaptive} method for setting the importance weights.  This modifies the weights such that the objective optimizes an implicit mass-covering $f$-divergence.  Due to space constraints, we refer the reader to \citet{wang2018tailadaptive} for more details.  Other strategies for modifying the weights could be employed \citep{ionides2008truncated, vehtari2015pareto}, but the tail-adaptive method is easy to implement, preserving the simplicity of dropout.  Specifically, the weights are set according to the rank of each sample: \begin{equation}\label{eq:ta}
    \gamma_{s} = \frac{S}{\sum_{k} \mathds{1}[\tilde{w}_{k} \ge \tilde{w}_{s}]}, \ \ \  w_{s} = \frac{\gamma_{s}}{\sum_{k} \gamma_{k}}
\end{equation} where $\tilde{w}_{s}$ is the likelihood under the $s$th sample as defined in Equation \ref{iw_grad}.  These weights do not depend on the precise value of the likelihood and are only a function of the sample size $S$.  

We visualize the importance weights produced by each method---the lower bound ($w \propto 1$), importance weighted ($w \propto p(\vy|\cdot)$), and tail-adaptive ($w \propto \gamma_{s}$)---in Figure \ref{fig:is_dist} for an experiment on the \texttt{Energy} UCI data set.  The gray histogram shows the importance weights observed when using \citet{gal2016dropout}'s dropout rate: $p=.005$, $10$ samples.  We see that the distribution is strongly peaked at $1/10=.1$, i.e. the uniform weight used by the lower bound.  The blue histogram shows when the dropout rate is increased to $p=.5$.  Now there is less of a mode at $.1$, but optimization does not converge under this dropout setting without a significant increase in the number of hidden units.  Hence, there is a fragile interdependence between the noise parameters, the network architecture, and the distribution of the importance weights.  The red histogram shows the weight distribution for the tail-adaptive method.  Even though the dropout rate is $p=.005$, there is still diversity in the weights, speaking to the method's ability to better explore the posterior.  

\subsection{Light-Weight Inference via Variational EM}\label{subsec:em}  
Ideally we would like to go beyond MAP point estimates and obtain some form of posterior \emph{distribution}.  Unfortunately, even approximate inference for scale mixture priors can be challenging---especially when the hyper-priors are half-Cauchy or log-uniform.  Previous work performing variational inference for these and similar priors has had to incorporate truncated approximations \citep{neklyudov2017structured}, auxiliary variables \citep{ghosh2018structured, louizos2017bayesian}, non-centered parametrizations \citep{ingraham2017variational, ghosh2018structured, louizos2017bayesian}, and quasi-divergences \citep{hron2018dropout} for the sake of tractability.  To preserve the simplicity of dropout, we propose a light-weight inference procedure derived through \textit{variational expectation-maximization} (EM) \citep{beal2003variationalEM}.  For most variational Bayesian NN implementations, using variational EM with ADD or ARD priors can be implemented in a few lines of code---possibly in as little as one.  Below we detail the inference procedure for ADD, leaving the description for the ARD-ADD prior to supplementary materials.  

Following \citet{wu2018fixingVB}, we assume the posterior approximation $p(\mW, \tau | \vy, \mX) \approx q(\mW; \vphi) q(\tau) = \text{N}(\mW; \vmu_{\vphi}, \text{diag}\{\mSigma_{\vphi}\}) \delta[\bar{\tau}_{l}]$ where $\vphi=\{ \vmu_{\vphi}, \text{diag}\{\mSigma_{\vphi}\} \}$ and $\bar{\tau}_{l}$ are the variational parameters.  We assume $\delta[\bar{\tau}_{l}]$ is a \textit{pseudo-Dirac delta} \citep{nakajima2014analysis} so that the distribution has finite entropy.  The evidence lower bound (ELBO) for this approximation is \begin{equation}\begin{split}\label{em_elbo}
    & \log  p(\vy | \mX ) \ge \mathbb{E}_{q(\mW)}\left[ \log p(\vy | \mX, \mW ) \right] \\ &  \ \ - \mathbb{E}_{q(\tau)}\text{KL}\left[ q(\mW; \vphi) || p(\mW | \tau) \right]  - \text{KL}\left[ q(\tau) || p(\tau) \right] \\ &= \mathbb{E}_{\text{N}(\mW)}\left[ \log p(\vy | \mX, \mW ) \right] \\ &  \ \ - \text{KL}\left[ \text{N}(\mW; \vphi) || \text{N}(\mW | \bar{\tau}_{l}) \right] - \text{KL}\left[ \delta[\bar{\tau}_{l}] || p(\tau) \right] \\ &= \mathbb{E}_{\text{N}(\mW)}\left[ \log p(\vy | \mX, \mW ) \right] \\ &  \ \ - \text{KL}\left[ \text{N}(\mW; \vphi) || \text{N}(\mW | \bar{\tau}_{l}) \right]  + \log p(\bar{\tau}_{l})  + \mathcal{C},
\end{split}\end{equation} where $\mathcal{C} = \mathbb{H}[\delta[\bar{\tau}_{l}]]$ is a constant.  For inverse-gamma, half-Cauchy, and log-uniform hyper-priors (and possibly others), $\bar{\tau}_{l}$ has a closed-form solution that can be found by differentiating the ELBO (Equation \ref{em_elbo}) and setting to zero. 
We denote the optimal solution 
as $\bar{\tau}_{l}^{*}$ and give its formula for each hyper-prior in the supplementary materials.  No closed-form exists for updating $q(\mW; \vphi)$, and hence we perform gradient ascent updates using \begin{equation}\begin{split}
   \frac{\partial}{\partial \vphi} & \mathcal{J}_{\text{\tiny{ELBO}}} ( \vphi, \bar{\tau}^{*}_{l}) =  \frac{\partial}{\partial \vphi} \mathbb{E}_{\text{N}(\mW; \vphi)}\left[ \log p(\vy | \mX, \mW ) \right] \\  & \ \ \ \ \ \ \ \ \ \ \ \ \ \ \ \ \ \ \ \ \ \ - \frac{\partial}{\partial \vphi} \text{KL}\left[ \text{N}(\mW; \vphi) || \text{N}(\mW | \bar{\tau}_{l}^{*}) \right]. 
\end{split}\end{equation}  
Figure \ref{fig:compare_w_vs} (b) shows the value for $\bar{\tau}_{l}^{*}$ as a function of the variational parameters $\vmu_{\vphi}$ and $\vsigma_{\vphi}^{2}$.  The slope and intercept of each line convey the prior's shrinkage properties.  Only the log-uniform and half-Cauchy provide true sparsity, allowing for $\bar{\tau}^{*}_{l} = 0$ when $\vmu_{\vphi}^{2} + \vsigma_{\vphi}^{2}=0$, no matter the setting of the prior's scale.  The inverse Gamma, on the other hand, can set $\bar{\tau}^{*}_{l} = 0$ only in the limit as $\alpha \rightarrow 0$ and $\beta \rightarrow 0$.   

\section{Related Work}
\citet{gal2016dropout, gal2016theoretically}'s interpretation of dropout as a variational approximation is perhaps the best known work contextualizing dropout within the Bayesian paradigm.  Their variational model is a spike-and-slab distribution and thus is equivalent to our generative model when $p(\xi)$ is Bernoulli.  However, there are crucial differences between their formulation and ours.  Firstly, their framework does not separate the model from inference, providing no direction on how one could employ dropout if performing inference via MCMC, for instance.  Since we work in terms of priors, MCMC can be applied as usual.  Secondly, \citet{gal2016dropout, gal2016theoretically} define their variational model as having NN weights $\mW = \mathbf{M} \ \text{diag}[\mathbf{z}]$ where $\mathbf{M}$ is a Gaussian matrix and $\text{diag}[\mathbf{z}]$ is a diagonal matrix with Bernoulli variables drawn from a \emph{fixed} distribution.  As far as we are aware, there is no reason for why the noise distribution must remain fixed, and in follow-up work \citet{gal2017concrete} relax this restriction, which had also been explored by \citet{ba2013adaptive}, \citet{wang2013fast}, \citet{maeda2014bayesian}, \citet{2016arXiv161106791S}, and \citet{zhai2018adaptive}.  Our work, on the other hand, derives their variational approximation from the perspective of structured priors.  Since the noise distribution is considered \emph{a prior}, it is natural that the dropout probability remains fixed and not optimized.  Consequently, our framework withstands the criticism that variational dropout's posterior does not contract as more data arrives \citep{osband2016risk}.

\citet{kingma2015variational} also propose a variational interpretation of dropout, which again couples the model with the inference strategy.  Their approach is derived by reparametrizing noise on the weights as uncertainty in the hidden units.  They show that their variational framework implies a log-uniform prior: $p(w) \propto 1/|w|$.  While our proposed GSM priors do not exactly match \citet{kingma2015variational}'s prior, the log-uniform does have heavy tails and strong shrinkage behavior near the origin, and interestingly, many of the marginal priors in the GSM family such as the student-t and horseshoe have those same characteristics.  However, recent work by \citet{hron2018dropout} illuminates flaws in the KL divergence term derived by \citet{kingma2015variational} for their implicit prior---specifically, that it results in an improper posterior.  Again, as our framework is removed from the variational paradigm and uses well-studied priors, this criticism does not apply.  \citet{molchanov2017variational} points out that \citet{kingma2015variational}'s dropout has the ARD structure, but they do not consider how this structure is derived from MN a priori nor ARD's extension to other architectures, as we do. 

Not all previous work has assumed a variational interpretation.  General treatments of data and/or parameter corruption as a Bayesian prior were considered by \citet{herlau2004bayesian} and \citet{nalisnick2018learning}.  The connection between dropout and the spike-and-slab prior has been noted by several previous works \citep{louizos2015smart, mohamed2015statistical, ingraham2017variational, 2018arXiv180309138P}.  \citet{ingraham2017variational} even note that dropout corresponds to "scale noise," but their primary focus was inference for undirected graphical models, not NNs.  Our work is distinct from these approaches in that none of them explicitly showed the equivalence via \citet{kuo1998variable}'s expanded parametrization, discussed other scale distributions, performed analysis of the MN objective, or generalized the interpretation to other architectures.

Regarding the inference algorithms discussed in Section \ref{sec:inf}, \citet{noh2017regularizing} recognized that the importance weighted objective in Equation \ref{eq:is_obj}, which was first proposed by \citet{burda2015importance} for variational inference, would provide a better estimator of the noise-marginalized likelihood.  However, they do not draw any connections to Bayesian inference.  \citet{wang2018tailadaptive}'s tail-adaptive method was proposed for general variational inference and its application to MN regularization has not been previously investigated.  The variational EM algorithm described in Section \ref{subsec:em} was inspired by \citet{wu2018fixingVB}'s ``empirical Bayes'' procedure.\footnote{We do not follow \citet{wu2018fixingVB}'s ``empirical Bayes'' naming convention because the hyperprior's parameters remain fixed and are not chosen by the data \citep{mackay1992bayesian}.}  We have extended their framework by considering structured priors and deriving the scale updates for additional hyperpriors (e.g.\ half-Cauchy and log-uniform).  

\begin{table*}
\centering
\resizebox{.99\linewidth}{!}{%
\begin{tabular}{l  c | c | c || c | c | c  }
& \multicolumn{3}{c}{\textbf{Test Set RMSE}} & \multicolumn{3}{c}{\textbf{Test Log-Likelihood}} \\
 & Lower Bound  & Import.\ Weighted &  Tail-Adaptive & Lower Bound  & Import.\ Weighted &  Tail-Adaptive  \\
\hline  
\texttt{Boston} & $2.80 \ \scriptstyle{\pm .19}$ & $ 2.450 \ \scriptstyle{\pm .25 }$ & $\mathbf{2.369 \ \scriptstyle{\pm .22}}$ & $-2.39 \ \scriptstyle{\pm .05}$ & $ -2.352 \ \scriptstyle{\pm .10 }$ & $ \mathbf{-2.346 \ \scriptstyle{\pm .01}}$ \\
\texttt{Concrete} & $4.81 \ \scriptstyle{\pm .14}$ & $4.052 \ \scriptstyle{\pm .29}$ & $\mathbf{3.935 \ \scriptstyle{\pm .35}}$ & $ -2.94 \ \scriptstyle{\pm .02 }$ & $ \mathbf{-2.888 \ \scriptstyle{\pm .08} }$ & $ -2.940 \ \scriptstyle{\pm .13}$ \\
\texttt{Energy} &  $1.09 \ \scriptstyle{\pm .05}$ & $0.972 \ \scriptstyle{\pm .06}$ & $\mathbf{0.828 \ \scriptstyle{\pm .05}}$  & $-1.72 \ \scriptstyle{\pm .02}$ & $\mathbf{-1.339 \ \scriptstyle{\pm .06}}$ & $-1.349 \ \scriptstyle{\pm .04}$  \\ 
\texttt{Kin8nm} & $0.09 \ \scriptstyle{\pm .00}$ & $0.082 \ \scriptstyle{\pm .00}$ & $\mathbf{0.076 \ \scriptstyle{\pm .00}}$ & $0.97 \ \scriptstyle{\pm .01}$ & $\mathbf{1.119 \ \scriptstyle{\pm .03}}$ & $1.105 \ \scriptstyle{\pm .03}$ \\
\texttt{Power} & $4.00 \ \scriptstyle{\pm .04}$ & $ \mathbf{3.094 \ \scriptstyle{\pm .08} }$ & $ 3.286 \ \scriptstyle{\pm .08 }$ & $ -2.79 \ \scriptstyle{\pm .01 }$ & $ \mathbf{-2.775 \ \scriptstyle{\pm .04} }$ & $ -2.809 \ \scriptstyle{\pm .05} $ \\
\texttt{Wine} & $0.61 \ \scriptstyle{\pm .01}$ & $0.667 \ \scriptstyle{\pm .02}$ & $\mathbf{0.559 \ \scriptstyle{\pm .03}}$ & $\mathbf{-0.92 \ \scriptstyle{\pm .01}}$ & $-0.996 \ \scriptstyle{\pm .04}$ & $-0.962 \ \scriptstyle{\pm .07}$ \\
\texttt{Yacht} & $0.72 \ \scriptstyle{\pm .06}$ & $\mathbf{0.577 \ \scriptstyle{\pm .16}}$ & $0.612 \ \scriptstyle{\pm .14}$ & $-1.38 \ \scriptstyle{\pm .01}$ & $\mathbf{-1.274 \ \scriptstyle{\pm .12}}$ & $-1.290 \ \scriptstyle{\pm .11}$
\end{tabular}
}
\caption{\textit{Comparing Monte Carlo Objectives}.  We compare test set RMSE and test log-likelihood for UCI regression benchmarks.  \citet{gal2016dropout}'s results are reported in the \textit{lower bound} column.}
\label{table:ss}
\end{table*}

\begin{figure*}
\begin{minipage}{0.745\linewidth}
\centering
\resizebox{.99\linewidth}{1.95cm}{\begin{tabular}{l  c | c | c || c | c | c }
& \multicolumn{5}{c}{\textbf{Test Set RMSE}} \\
 & Dropout & Prob.\ Backprop & Deep GP &  ARD  & ADD  & ARD-ADD \\
\hline 
\texttt{Boston} & $2.80 \ \scriptstyle{\pm .13}$ & $2.795 \ \scriptstyle{\pm .16 }$ & $2.38 \ \scriptstyle{\pm .12 }$ & $\mathbf{2.158 \ \scriptstyle{\pm .20} }$ & $2.343 \ \scriptstyle{\pm .31  }$ & $2.367 \ \scriptstyle{\pm .18}$ \\
\texttt{Concrete} & $4.50 \ \scriptstyle{\pm .18}$ & $5.241 \ \scriptstyle{\pm .12}$  & $4.64 \ \scriptstyle{\pm .11}$ & $3.805 \ \scriptstyle{\pm .28}$ &   $4.084 \ \scriptstyle{\pm .34}$  & $\mathbf{3.761 \ \scriptstyle{\pm .23}}$ \\
\texttt{Energy} &  $\mathbf{0.47 \ \scriptstyle{\pm .01}}$ &  $0.903 \ \scriptstyle{\pm .05}$ & $0.57 \ \scriptstyle{\pm .02}$ & $0.852 \ \scriptstyle{\pm .01}$ & $0.867 \ \scriptstyle{\pm .11}$ & $0.853 \ \scriptstyle{\pm .08}$ \\ 
\texttt{Kin8nm} & $0.08 \ \scriptstyle{\pm .00}$ & $0.071 \ \scriptstyle{\pm .00}$  & $\mathbf{0.05 \ \scriptstyle{\pm .00}}$ & $0.066 \ \scriptstyle{\pm .01}$ &   $0.064 \ \scriptstyle{\pm .00}$  & $0.064 \ \scriptstyle{\pm .00}$ \\
\texttt{Power} & $ 3.63 \ \scriptstyle{\pm .04}$ & $4.028 \ \scriptstyle{\pm .03} $  & $3.60 \ \scriptstyle{\pm .03}$ & $3.486 \ \scriptstyle{\pm .10}$ &   $3.290 \ \scriptstyle{\pm .06}$  & $\mathbf{3.236 \ \scriptstyle{\pm .07}}$ \\
\texttt{Wine} & $ 0.60 \ \scriptstyle{\pm .01}$ & $0.643 \ \scriptstyle{\pm .01}$  & $\mathbf{0.50 \ \scriptstyle{\pm .01}}$ & $0.561 \ \scriptstyle{\pm .03}$ &   $0.555 \ \scriptstyle{\pm .01}$  & $0.538 \ \scriptstyle{\pm .03}$ \\
\texttt{Yacht} & $0.66 \ \scriptstyle{\pm .06}$ & $0.848 \ \scriptstyle{\pm .05}$ & $0.98 \ \scriptstyle{\pm .09}$ & $0.691 \ \scriptstyle{\pm .12}$ & $0.657 \ \scriptstyle{\pm .14}$ & $\mathbf{0.604 \ \scriptstyle{\pm .16}}$ \\ \hline Avg.\ Rank & $4.4 \ \scriptstyle{\pm 1.7}$ & $5.6 \ \scriptstyle{\pm 0.5}$ & $3.1 \ \scriptstyle{\pm 1.8}$ &  $3.0 \ \scriptstyle{\pm 1.1}$ & $2.9 \ \scriptstyle{\pm 10}$ & $\mathbf{2.0 \ \scriptstyle{\pm 1.1}}$  
\end{tabular}}
\captionsetup{type=table}
\caption{We compare test set RMSE for UCI regression benchmarks.  As baselines, we use previously reported two-hidden-layer results for dropout \cite{gal2016dropout}, probabilistic backpropagation \cite{hernandez2015probabilistic}, and deep Gaussian processes \citep{pmlr-v48-bui16}.  ARD, ADD, and ARD-ADD use the $\Gamma^{-1}(3,3)$ scale prior in all cases.}
\label{fig: table-label}
\end{minipage}
\hfill
\begin{minipage}{0.25\linewidth}
\centering
\subfigure{\includegraphics[width=.99 \linewidth]{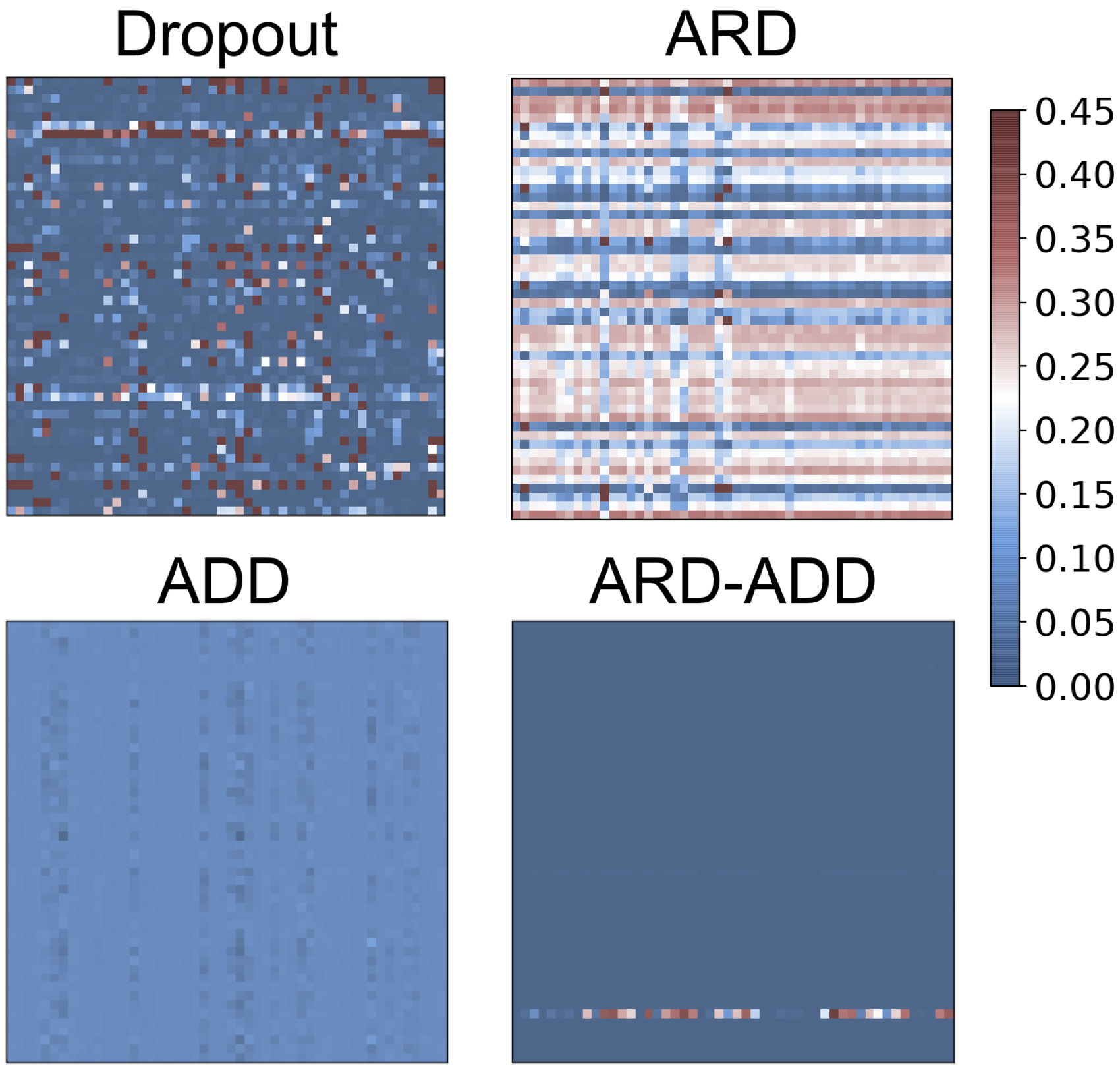}}
\caption{\textit{Posterior Structure.}}
\label{tab: fig-label}
\end{minipage}
\end{figure*}

\section{Experiments}
We performed experiments to test the practicality of the tail-adaptive importance sampling scheme (Section \ref{subsec:is}) for MC dropout and the variational EM algorithm (Section \ref{subsec:em}) for ARD, ADD, and ARD-ADD priors (Section \ref{sec:resnets}).  For both cases we used the same experimental setup as \citet{gal2016dropout}, testing the models on regression tasks from the UCI repository \citep{Dua:2017}.  The supplementary materials include details of the model and optimization hyperparameters as well as Python implementations\footnote{Available at: \url{https://github.com/enalisnick/dropout_icml2019}} of both experiments.

\paragraph{Tail-Adaptive Importance Sampling}  Table \ref{table:ss} reports test set root-mean-square error (RMSE) and log-likelihood for three MN objectives: the usual lower bound (Equation \ref{dropout_loss}), the importance weighted objective (Equation \ref{eq:is_obj}), and the tail-adaptive method (Equation \ref{eq:ta}).  The `lower bound' columns are the results reported by \citet{gal2016dropout}.\footnote{We report \citet{gal2016dropout}'s updated experiments that appear in Appendix Table 2 of the ArXiv version.}  The only difference between columns is in how the importance weights were set.  We see that the tail-adaptive method results in the best RMSE in five out of seven data sets.  However, the best log-likelihoods are achieved by regular importance sampling (five out of seven).  We believe that this is due to $\mathcal{L}_{\text{IW-MN}}$ being a less-biased estimate of the exact likelihood (unbiased as $S\rightarrow \infty$).  The tail-adaptive method, on the other hand, is most effective at regularizing for purposes of predictive accuracy.  Notably, the tail-adaptive method performs worst on \texttt{Power}, the largest data set ($N =  9568)$.

\paragraph{EM for Resnets} We next report results for the variational EM algorithm applied to resnets with two hidden layers (one skip connection) and ARD, ADD, and ARD-ADD priors.  We compare their test set RMSE with the two-layer results reported for dropout \citep{gal2016dropout}, probabilistic backpropagation \citep{hernandez2015probabilistic}, and deep Gaussian processes (GPs) trained with expectation propagation (EP) \citep{pmlr-v48-bui16}.   
Table \ref{fig: table-label} contains the results: the ARD-ADD resnet gives the best performance on three data sets, and the deep GP gives the best on two.  The average rank of each model / algorithm is given in the last line with ARD-ADD coming in first (but there is substantial overlap in the error bars).  This result is encouraging since the EM-trained ARD-ADD resnet can compete with the deep GP, a rich nonparametric model, trained with EP, a strong approximate inference algorithm \citep{epWayofLife}.  

Figure \ref{tab: fig-label} shows heat maps of the hidden-to-hidden weight matrices for dropout and the three ARD/ADD priors when trained on \texttt{Boston}.  The colors are determined by the summed posterior moments $\mu_{\vphi}^{2} + \sigma_{\vphi}^{2}$ (just $w^{2}$ for dropout) as this quantifies each parameter's ability to deviate from zero.  We see that, as expected, ARD induces row-structured shrinkage, ADD induces matrix-wide shrinkage, and ARD-ADD allows some rows to grow (just one in this case) while preserving global shrinkage.  While not strictly comparable to the others, MC dropout's heat map seems to balance having some row structure with strong global shrinkage.  

\section{Conclusions}
We have presented a novel interpretation of MN, showing that it induces an ARD-GSM prior.  This revelation uncouples dropout's generative assumptions from inference, unlike previously proposed Bayesian interpretations \citep{kingma2015variational, gal2016dropout}.  Our Bayesian framework then inspires an extension of ARD to resnets, a novel prior that we call \textit{automatic depth determination}, and the application of two alternative inference algorithms.  An exciting direction for future work is to extend the ARD framework to recurrent and convolutional networks.  Bayesian inference for these architectures has proven to be challenging \citep{gal2016conv, fortunato2017bayesian}, and the structured priors and efficient inference algorithms we explore in this work may enable a notable jump in progress.    

\newpage
\subsubsection*{Acknowledgements}
We thank Anima Anandkumar and Robert Peharz for helpful discussions.  E.N.\ and J.M.H.L.\ gratefully acknowledge support from Samsung Electronics.  

\bibliography{references}
\bibliographystyle{icml2019}

\end{document}